\documentclass{article}
\usepackage{ijcai26}

\usepackage{times}
\usepackage{soul}
\usepackage{url}
\usepackage[hidelinks]{hyperref}
\usepackage[utf8]{inputenc}
\usepackage[small]{caption}
\usepackage{graphicx}
\usepackage{amsmath}
\usepackage{booktabs}
\usepackage{algorithm}
\usepackage{algorithmic}
\usepackage[switch]{lineno}
\usepackage{dirtytalk}
\usepackage{float}

\title{Addressing the Selection Problem in Explainable AI}

\author{
Claire Vlases$^1$
\and
Katelyn Morrison$^1$\\
\affiliations
$^1$Carnegie Mellon University, Pittsburgh, PA, USA\\
\emails
\{cvlases, kcmorris\}@andrew.cmu.edu
}

\begin{document}

\maketitle

\begin{abstract}
    Explainable AI (XAI) research has produced a plethora of explanation
techniques, yet user studies repeatedly show that available explanations
are not effective in practice. We argue that, given the siloed nature of conventional XAI, users are struggling to select the appropriate XAI technique. Viewing XAI through a philosophical lens, we offer a formalization of what we call the \emph{selection problem}: the systematic failure of XAI interfaces to bridge the gap between a user's natural-language uncertainty and the explanation technique that resolves it. Following a logical premise-conclusion format, we show that conventional interfaces require users to translate their uncertainty into a technique selection, a challenging prerequisite to meet. We also propose a structural solution: a multi-agent LLM orchestration tool that translates the user's query to the proper XAI explanation technique. We provide an example of how this structural solution could be instantiated to address the selection problem.

\end{abstract}

\section{Introduction}

In recent years, the rapidly emerging field of explainable AI (XAI) has produced a diverse landscape of explanation techniques \cite{mersha-et-al:xai-survey,wilkinson-et-al:explaining-explainability,schwalbe_comprehensive_2023,schneider:genxai-survey}. Many such explanations are unimodal and non-interactive: the user has limited control over what is shown and limited ability to request follow-up information. Techniques are also often evaluated in isolation and compared against one another to identify the best configuration of a single method (\textit{e.g.},~\cite{humer_comparing_2022}). Studies have explored a variety of XAI techniques. For example, \textit{saliency} methods identify which input regions drive a prediction \cite{Selvaraju_2019}. \textit{Counterfactual} methods show how small input changes alter predictions and thereby reveal critical decision factors \cite{Mothilal_2020}. \textit{Exemplar} methods produce explanations by retrieving the closest examples from a training dataset \cite{hou-wang:clinical-xai}. \textit{Query-based} methods let users question model perception directly \cite{lakkaraju2026holisticexplainableaihxai}. Each technique answers a structurally different question about AI behavior.

However, the evaluation of these XAI techniques in practice with real users remains limited \cite{suh-et-al:fewer-than-1}. User studies show that available explanations are frequently ignored, misinterpreted, or used to confirm rather than scrutinize AI outputs, leading to an inappropriate reliance on AI explanations \cite{misinterpret-acm}. Some literature has attributed these challenges to the lack of human-centered techniques~\cite{mangold-et-al:human-centered-xai}. 

Recent work has begun exploring conversational and interactive forms of explanation, including narrative-driven XAI with large language models \cite{martens2026tellstorynarrativedrivenxai} and proposals for agentic interpretability pursued through dialogue with a generative model \cite{kim2025llmspursueagenticinterpretability}. These directions share our view that explanations are better treated as an interaction than as a static output \cite{rohlfing2020explanation}. Our contribution formalizes the challenge of operationalizing XAI techniques within an interactive system.
 
Before receiving any explanation when a user is collaborating with an AI system, that user must answer an implicit question: \textit{which technique is appropriate for what I currently want to understand?} Conventional XAI systems offer explanation techniques as discrete controls. However, providing individual explanation methods themselves is not sufficient for users to judge, trust, or question an AI effectively \cite{labarta-et-al:helpfulness-xai}. Using a philosophic research lens, we formalize this as the \textit{\textbf{selection problem in XAI}}: the failure of XAI interfaces to bridge the gap between a user's natural-language expression of uncertainty and the explanation technique that resolves it. We formalize the selection problem as a property of conventional XAI interface design that places the burden of mapping uncertainty to the proper explanation technique on the user when it could be handled by the system.
 
We then derive three requirements for a structural solution: it must accept natural-language input directly; output adequate explanation techniques; and perform the translation mapping on behalf of the user. Given that a user expresses uncertainty in natural language, we propose \textit{conversational multi-agent orchestration of XAI} as a possible design solution.
 

\section{The Selection Problem in XAI}
\label{sec:formalization}
 
We formalize the selection problem as a structural property of conventional XAI interface design. The argument proceeds in three stages: different epistemic states require different techniques; users must perform a translation from their natural language uncertainty to an XAI technique selection; and the user's mapping may be unreliable.

\begin{description}\itshape
\item[\upshape Premise 1a.] An AI model is a function from inputs to outputs.
\item[\upshape Premise 1b.] An explanation technique is a function that takes an input and a model and returns an explanation.
\item[\upshape Premise 1c.] Different explanation techniques may answer structurally different questions.
\item[\upshape Premise 1d.] A user's uncertainty has a specific epistemic state at any given moment.
\item[\upshape Conclusion 1.] Given a user's uncertainty, only a subset of available techniques can actually resolve it.
\item[\upshape Premise 2a.] Users express uncertainty in natural language, not as
technique selections.
\item[\upshape Premise 2b.] Conventional XAI interfaces require a technique selection as input.
\item[\upshape Conclusion 2.] The user must perform a translation from natural language to technique selection.
\item[\upshape Premise 3.] Reliable question-to-XAI type mapping is not a skill the user population can be assumed to have.
\item[\upshape Conclusion 3.] User-performed mapping is unreliable.
\item[\upshape Conclusion.] Different uncertainties require different techniques (C1), and
users cannot reliably translate their uncertainty into the right technique (C2, C3).
Therefore, there exists a selection problem in which the user's selected technique falls outside the adequate set.
\end{description}

\paragraph{Premise 1a. An AI model is a function from inputs to outputs.}

To view the selection problem through a philosophical lens, we first define the structure of an AI model as a function of inputs and outputs. An AI model $M$ takes in an input space $X$, the set of all possible inputs, and maps to an output space $Y$, which contains all possible predictions generated by the system. The model is therefore a function $M : X \to Y$. Under this formalization, $M$ takes an input $x \in X$ and returns a prediction $y \in Y$.
 
\paragraph{Premise 1b. An explanation technique is a function that takes an input and a model and returns an explanation.}
With that understanding, we use the following definition of XAI: ``\textit{a set of processes and methods that allows human users to comprehend and trust the results and output created by machine learning algorithms}''~\cite{ibm:xai}. An explanation is the specific output produced by the technique when run on a particular input and model. For example, gradient class activation mapping (Grad-CAM) outputs a saliency map \cite{Selvaraju_2019}; the Diverse Counterfactual Explanations (DiCE) algorithm outputs a counterfactual \cite{Mothilal_2020}; nearest-neighbors retrieval outputs an example-based explanation \cite{dong2025obtainingexamplebasedexplanationsdeep}.
 
\paragraph{Premise 1c. Different explanation techniques answer structurally different questions.} 
Each technique produces a different kind of explanation output \cite{liao-et-al:questioning-ai}. We define a finite set of techniques as $E$, where any individual technique $e_i$ is an element of the set, $e_i \in E$. Each technique $e_i$ takes in input $x$ and model $M$ to produce an explanation.
 
\paragraph{Premise 1d. A user's uncertainty has a specific epistemic state at any given moment.}
A user seeks an XAI explanation when they encounter an AI prediction that leaves them uncertain about the model's logic; they need help understanding why $M$ produced prediction $y$ for input $x$. We represent the specific uncertainty as their epistemic state $\omega$. Because the epistemic state is conditioned upon the existing knowledge the user holds at the time of the interaction, we write $\omega = (x, y, \phi)$: the input to the model $x$, the output of the model $y$, and $\phi$, the specific gap in their understanding that prompts them to seek an explanation.

The uncertainty gap $\phi$ encompasses what the user does not understand but also why they seek to understand it. The goal of an explanation shapes which technique is adequate: a user seeking recourse is served by counterfactuals that modify a small number of features, while a user seeking to learn about a domain may be better served by more global explanations \cite{delaney-et-al:counterfactual-misclassified}. Two users with the same input, output, and surface question may have different reasons for wanting an explanation, thus possessing different epistemic states.
 
\paragraph{Conclusion 1. Only a subset of available techniques can resolve any given epistemic state.}
Because each explainability method is designed to generate a distinct type of output, its capacity to address the user's particular inquiry is necessarily constrained. The utility of any given technique is siloed, relevant only to a narrow set of questions. Within the context of an epistemic state $\omega$, only some of the techniques in $E$ are actually useful. We denote the subset of techniques capable of resolving the user's uncertainty as $E^*(\omega)$, where $E^*(\omega) \subseteq E$. This is the subset of explanations that,
given the user's epistemic state $\omega$, would actually help the user with what they want to know. The efficacy of an explanation technique is defined relative to the user's epistemic state.
 
\paragraph{Premise 2a. Users express uncertainty in natural language, not as technique selections.}
The epistemic state of the user $\omega$ is an internal condition that must be communicated externally to the XAI system. The user must convert their sense of ambiguity into a structured natural-language form, denoted as query $q$. The query $q$ serves as the only external evidence of the user's requirements. The user's underlying epistemic state is not directly accessible by the system, which can only perceive a representation of the uncertainty. The user articulates their internal uncertainty $\phi$ into a query $q \in L$, where $L$ represents the set of all possible expressions a user might say or think. The query $q$ functions as a verbalization of the underlying uncertainty $\phi$. 

Notably, $q$ is not assumed to be an accurate or complete representation of the underlying uncertainty $\phi$. A user's selection of an inadequate technique can itself signal unresolved uncertainty, but conventional interfaces have no channel through which to detect or act on that signal.
 
\paragraph{Premise 2b. Conventional XAI interfaces require a technique selection as input.}
Conventional XAI interfaces do not accept $q$ as input. In conventional systems, either the user selects the technique or the developer fixes the choice in advance on their behalf. Either way, a human agent (which we will refer to as the user, for convenience) must produce some chosen technique $\hat{e} \in E$.
 
\paragraph{Conclusion 2. The user must translate from natural language to technique selection.} The user must identify which methodology within the available set is capable of resolving their particular question, mapping their query $q$ to a technique choice $\hat{e}$. For this translation step, the user must possess enough technical knowledge to understand the capabilities of the various explanation techniques. Because both the articulation and the mapping phases occur on the user side of the interaction, the system is unable to detect if a failure occurs in either step. The system never sees the user's internal uncertainty, only their selected technique. The system can instead only access $q$, a representation of $\phi$. A mismatch between the user's curiosity and the system's output can persist without any detection by the model.
 

\paragraph{Premise 3. Reliable question-type classification is not a skill the user population can be assumed to have.}

One of the most referenced works on XAI is the XAI Question Bank (XAIQB), an algorithm-informed XAI question bank in which user needs for explainability are represented as typical questions users might ask about the AI \cite{liao-et-al:questioning-ai}. Users ``\textit{may hold preconceptions of what constitutes useful explanations for decisions},'' even without a deep technical understanding of AI \cite{liao-et-al:questioning-ai}. The XAIQB exists precisely because users cannot be expected to produce a categorization or mapping of XAI techniques on their own. Empirical applications of the XAIQB corroborate that some questions were difficult to distinguish and required interpretation during use, thus unclear to what extent the XAIQB captured the explanation needs of end-users \cite{sipos-et-al:explanation-needs}.

In practice, a user-performed mapping of an uncertainty to a explanation technique produces structured and recurring failure modes. A typology of XAI failures identifies user-side failures arising from inconsistent inferences, distinguishing mismatch failures (contradictions between ML explanations and users' expectations in terms of format), counterintuitive failures (contradictions in explanation content), and biased inferences (cognitive biases interfering with the explanation) \cite{bove2025explanationsfailtypologydiscussion}. Each of these failure modes is consistent with a user who has attempted to seek an explanation without a reliable mapping from their question to a technique type. The user receives an explanation, but it does not align with what they were trying to understand.

\paragraph{Conclusion 3. User-performed mapping is unreliable.}
It follows that for many users, their selection of an explanation technique $\hat{e}$ is not reliably constrained to fall within the adequate set $E^*(\omega)$. The user is forced to map without the prerequisites needed to perform that mapping reliably. This is a structural consequence of placing the mapping function $f_\textit{user}$ at a point in the workflow where the user is required to act before the system can support them.
 
\paragraph{Conclusion: There exists a \upshape\textit{selection problem} in which the user's selected technique falls outside the adequate set.}
We can write the user's translation as a function $f_\textit{user} : L \to E$ that turns a natural-language query into a technique selection. The user is forced to perform this mapping without any support from the system, and doing it correctly requires knowing what each tool does, which question types each tool answers, and being able to accurately name what kind of question they are asking, all before seeing any explanation at all. This prerequisite exists on both the user and developer sides. While developers may understand the operational parameters of the available explanation techniques, they may lack insight into the user's specific intent or the nature of their underlying question. End users may recognize their own uncertainty but lack the technical knowledge to identify the explanation tools that would resolve it.
 
When the user fails to map their uncertainty to the appropriate technique, and the selected explanation $\hat{e}$ lands outside of $E^*(\omega)$, we have what we call the \textit{selection problem} of XAI: the technique the user chose is not among the techniques that would actually resolve their uncertainty. More formally, the selection problem occurs when 
\begin{equation}
    \hat{e} \notin E^*(\omega).
    \label{eq:selection-problem}
\end{equation}
The selection problem represents a failure in the explanatory process, where the output provided by the system remains functionally irrelevant to the user's actual needs.
 
\section{Structural Solution: Multi-Agent LLM Orchestration System}

The selection problem arises because conventional XAI interfaces require the user to perform the mapping $f_\textit{user} : L \to E$ from query to explanation technique. The structural solution is to shift this burden to the system, a mapping function $f_\textit{sys}$.
 
We identify three requirements for any system that replaces user-performed mapping: it must accept natural-language input, it must return adequate techniques as output, and the translation between them must be performed by the system rather than the user.
 
\textbf{The system accepts natural-language input.} The user's uncertainty originates in natural language $L$. Conventional interfaces demand a discrete technique selection, or they force a selection by only providing a single technique. The mismatch between these two forces the user to translate their query into a vocabulary the interface accepts. A structural solution must therefore accept $q \in L$ directly, so the user is not required to map their question to technique-based terms. 
 
\textbf{The system returns adequate technique routing.}
Each explanation technique answers a distinct question, and there is an adequate technique set $E^*(\omega) \subseteq E$ for each epistemic state. A solution that returned techniques outside $E^*(\omega)$ would deliver explanations that fail to address the user's actual uncertainty. The system must therefore aim to select techniques that can resolve the user's expressed uncertainty.


\textbf{The system assists with the mapping translation.}
One solution could move the mapping function off the user partially or entirely, so that $f_\textit{sys} : L \to E^*(\omega)$ is performed by the system. The user's remaining contribution is to articulate $q$. Although the user's articulation itself may not be trivial, shifting the mapping to the system opens the possibility of a collaborative, multi-turn exchange. The system could respond to internal uncertainty in ways a conventional interface cannot, such as eliciting the user's goal or refining $q$ before routing.

\section{Instantiating a Multi-Agent LLM Orchestrator for XAI}
 
We provide the community with an example of how to instantiate this structural solution using four different XAI techniques in the context of medical imaging. While we believe this structural solution can be domain-agnostic, describing its potential operationalization in a specific domain helps illustrate the functionality. 

  \begin{figure}[!ht]
    \centering
    \includegraphics[width=0.27 \textwidth]{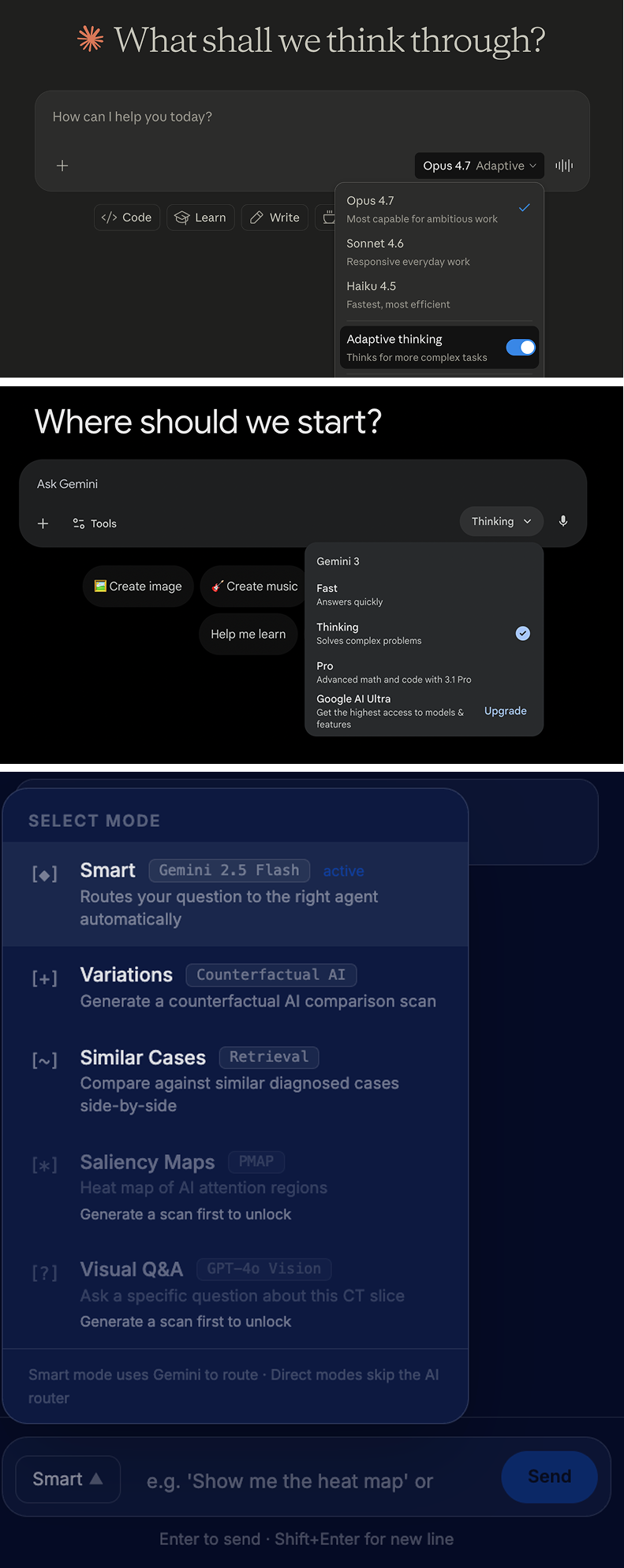}
    \caption{(Top \& Middle) Examples of existing public LLM interfaces with multi-agent functionality used for design inspiration. (Bottom) Example design of offering an LLM orchestrator ``Smart'' mode that routes natural-language queries from the user to one of the specialized XAI agents through function calling. }
    \label{fig:architecture}
\end{figure}

The first system requirement demands natural-language understanding at the input boundary. The second requires knowledge of which techniques resolve which kinds of uncertainty, encoded in a way that supports automated routing. The third calls for a system-run process that routes a natural-language query input to an explanation technique as output.

Drawing from designs in popular public LLM tools (top two images of Figure~\ref{fig:architecture}), such as Claude, Gemini, and ChatGPT, we imagine how a multi-agent LLM orchestrator could help users navigate different XAI techniques as shown with the ``Smart'' mode in the bottom image of Figure~\ref{fig:architecture}. A ``Smart'' orchestration tool offers a blueprint: prompt an LLM to match the user's natural-language question to the question types that known XAI techniques address, such as those in Liao's XAIQB. Routing decisions are made by the orchestrator at the moment of query, ensuring that the techniques returned are matched to the structure of the user's uncertainty.

The orchestrator tool would thus model $f_\textit{sys} : L \to E^*(\omega)$, receiving the user's natural language query, classifying its underlying type, and routing it to the best-suited explanation agent. For users who already know which explanation technique they want, individual ''modes'' can be invoked directly, bypassing the orchestrator. In our instantiation, the orchestrator itself is implemented with Gemini 2.5 Flash, configured to map query types onto the available agents.

Our LLM-based orchestrator instantiation aims to address the three requirements by mapping the user's query to a suite of XAI agents. Natural-language understanding is a native LLM capability, satisfying the first requirement. Function calling provides the routing mechanism, satisfying the second. The ability of the orchestrator to interpret a query, select techniques, and return outputs in a single pipeline absorbs the mapping function into the system, satisfying the third.

\section{Future Work \& Conclusion}
 
We have argued that the adoption gap in XAI is, at least in part, a problem of interface presentation. Conventional XAI interfaces force the user to map their natural-language uncertainty onto a discrete technique selection before any explanation is produced, and performing this mapping can be unreliable. We formalized this as the \textit{selection problem} and proposed an LLM-based multi-agent orchestrator that absorbs the mapping function into the system as one solution. 

Future work should consider the selection problem more closely by further investigating the premises and conclusions we present. For example, researchers could design studies to more concretely identify whether user-performed mapping is indeed unreliable and to what extent the selection problem exists for end-users versus AI system designers. 

Beyond that, researchers could expand these designs to other domains and XAI techniques. An immediate next step could be to compare the effectiveness of a multi-agent LLM orchestrator for XAI with that of presenting the techniques to users without the orchestrator. Technical researchers could explore how to best implement the backend of the orchestrator to ensure that the user's natural language inquiries are routed to the correct XAI agent, whether through prompt engineering or by implementing multi-turn conversations. 

Ultimately, we hope that our formalization, description of system requirements, and example implementation of the system inspire conversations in the human-centered XAI community and encourage researchers to consider the selection problem when designing XAI systems.

\section*{Acknowledgments}

\textbf{Use of Generative AI Disclosure.} Claude was used during manuscript preparation to explore and refine candidate formalizations, notation, and descriptions. The authors selected, revised, extended, and verified the final formalization and all claims/positions in the paper.

 \clearpage
\bibliographystyle{named}
\bibliography{ijcai26}

\end{document}